\documentclass[11pt]{article}

\usepackage[final]{acl}

\usepackage{times}
\usepackage{latexsym}
\usepackage{hyperref}
\usepackage{booktabs}    
\usepackage{multirow}    

\usepackage[T1]{fontenc}

\usepackage[utf8]{inputenc}

\usepackage{microtype}

\usepackage{inconsolata}

\usepackage{graphicx}

\title{KazakhOCR: A Synthetic Benchmark for Evaluating Multimodal Models in Low-Resource Kazakh Script OCR}

\author{
  \textbf{Henry Gagnier\textsuperscript{1}},
  \textbf{Sophie Gagnier\textsuperscript{1}},
  \textbf{Ashwin Kirubakaran\textsuperscript{2}}
\\
  \textsuperscript{1}Pittsford Sutherland High School,
  \textsuperscript{2}Edison Academy Magnet School,
\\
  \small{
    \textbf{Correspondence:} \href{mailto:henrygagnier9@gmail.com}{henrygagnier9@gmail.com}
  }
}

\begin{document}
\maketitle
\begin{abstract}
Kazakh is a Turkic language using the Arabic, Cyrillic, and Latin scripts, making it unique in terms of optical character recognition (OCR). Work on OCR for low-resource Kazakh scripts is very scarce, and no OCR benchmarks or images exist for the Arabic and Latin scripts. We construct a synthetic OCR dataset of 7,219 images for all three scripts with font, color, and noise variations to imitate real OCR tasks. We evaluated three multimodal large language models (MLLMs) on a subset of the benchmark for OCR and language identification: Gemma-3-12B-it, Qwen2.5-VL-7B-Instruct, and Llama-3.2-11B-Vision-Instruct. All models are unsuccessful with Latin and Arabic script OCR, and fail to recognize the Arabic script as Kazakh text, misclassifying it as Arabic, Farsi, and Kurdish. We further compare MLLMs with a classical OCR baseline and find that while traditional OCR has lower character error rates, MLLMs fail to match this performance. These findings show significant gaps in current MLLM capabilities to process low-resource Abjad-based scripts and demonstrate the need for inclusive models and benchmarks supporting low-resource scripts and languages.
\end{abstract}
\section{Introduction}
Kazakh is a Turkic language spoken in Kazakhstan, China, Mongolia, Russia, Kyrgyzstan, and Uzbekistan by 16 million people \cite{McCollum_Chen_2020}. Kazakh is written in different scripts depending on the geographic and political context. In Central Asia, a Cyrillic script is used for Kazakh, while in Europe, America, and Turkey, a Latin script is used, and in China, Afghanistan, Pakistan, and Iran, an Arabic script is used \cite{honkasalo_temirbekova_2024}. The use of these different writing scripts makes Kazakh unique and makes optical character recognition (OCR) particularly complex, considering Cyrillic, Latin, and Arabic scripts.

OCR refers to the electronic translation of handwritten, typewritten, or printed text, and its conversion into a machine-readable form \cite{Faizullah_Ayub_Hussain_Khan_2023}. It enables the recognition of text in digital images, scanned documents, and videos, converting images into machine-coded text. OCR has many applications, including document scanning and form processing. Convolutional neural networks (CNNs) and transformer architectures have recently improved OCR performance across many scripts and languages. Multimodal models have been used for OCR, enabling text extraction without segmentation and allowing for greater robustness with noise. Research on these models has primarily focused on high-resource languages and has not been performed with the Kazakh Arabic script.

Prior work for Kazakh OCR has been primarily done on the Cyrillic script. \citet{Toiganbayeva_Kasem_Abdimanap_Bostanbekov_Abdallah_Alimova_Nurseitov_2022} created a KOTHD, a Kazakh offline handwritten text dataset using Cyrillic script. \citet{Razaque_Makezhanuly_Alimseitov_Kalpeyeva_Ayapbergenova_2024} developed a handwritten text recognition system for handwritten Cyrillic script using CNNs and RNNs. \citet{Pirniyazova_Son_Kulzhan_2023} evaluates a neural network on the recognition of Latin letters of the Kazakh alphabet. \citet{Yeleussinov_Amirgaliyev_Cherikbayeva_2023} use a Generative Adversarial Network (GAN) to improve Kazakh handwriting recognition using Cyrillic script. The Kazakh Arabic script uses additional letters, modified graphemes, and different orthographic conventions, making it different from high-resource Arabic varieties and an important test case for Abjad-based low-resource NLP. While work has been performed on Kazakh OCR, it does not focus on the Arabic script and scarcely focuses on the Latin script.

Synthetic datasets are emerging as a solution to a lack of images and content in low-resource languages. The creation of a synthetic OCR dataset often consists of the collection of text corpora, fonts, and noise filters \cite{saini2022ocrsyntheticbenchmarkdataset}. \citet{Haq_Zhang_Khan_2025} developed a synthetic benchmark for Pashto and the Pashto script. \citet{953967} proposed a method for the creation of synthetic data for Arabic OCR. \citet{saini2022ocrsyntheticbenchmarkdataset} synthetically created a benchmark of 90k images for 23 Indic languages. Synthetic data has also been used for post-OCR correction, and has been shown to have advantages over traditional OCR training in low-resource languages \cite{guan2024advancingpostocrcorrectioncomparative}. Synthetic data is particularly useful when limited resources for OCR are available, such as in low-resource Kazakh scripts.

Research on OCR for Kazakh using the Arabic script is scarce, but extremely necessary. The purpose of this study is to (1) construct synthetic OCR benchmarks for low-resource Kazakh scripts, (2) evaluate multimodal models on the OCR of low-resource Kazakh scripts, and (3) identify challenges in the OCR of Kazakh. This study also aims to encourage the inclusion and research of low-resource scripts, particularly languages using the Arabic Kazakh script.
\section{Methodology}
\subsection{Kazakh Text Corpora}
While it is possible to obtain perfect transliterations in Kazakh using pre-defined rules, cultural differences in the Cyrillic and Arabic scripts make transliteration
from high-resource scripts into low-resource scripts sub-optimal for training \cite{zhang-etal-2024-mc2}. We obtained authentic Kazakh text in the Arabic script from the first Kazakh corpus in the Arabic script, the MC² corpus \citet{zhang-etal-2024-mc2}. Wikipedia is also a good resource for low-resource languages such as Kazakh. For Cyrillic text, we obtained authentic data from the Kazakh Wikipedia\footnote{\href{https://huggingface.co/datasets/wikimedia/wikipedia}{https://huggingface.co/datasets/wikimedia/wikipedia}}.

We cleaned the Cyrillic script data from Wikipedia using kaznlp \cite{10.1007/978-3-030-60276-5_63}, and transliterated it to the Latin script using the QazNLTK package\footnote{\href{https://pypi.org/project/qaznltk/}{https://pypi.org/project/qaznltk/}}. As the number of characters in each sample of the Arabic (3746 characters) script text was much greater than the number of characters in the Cyrillic (689 characters) and Latin (718 characters) script texts, we implemented random balanced subsampling to make the text lengths similar in each of the scripts. After balance subsampling, we had 2,417 Arabic texts, 2,402 Cyrillic texts, and 2,400 Latin texts, with similar text length distributions across scripts, which were all used in the final benchmark.
\subsection{Benchmark Construction}
\begin{figure}
    \centering
    \includegraphics[width=1\linewidth]{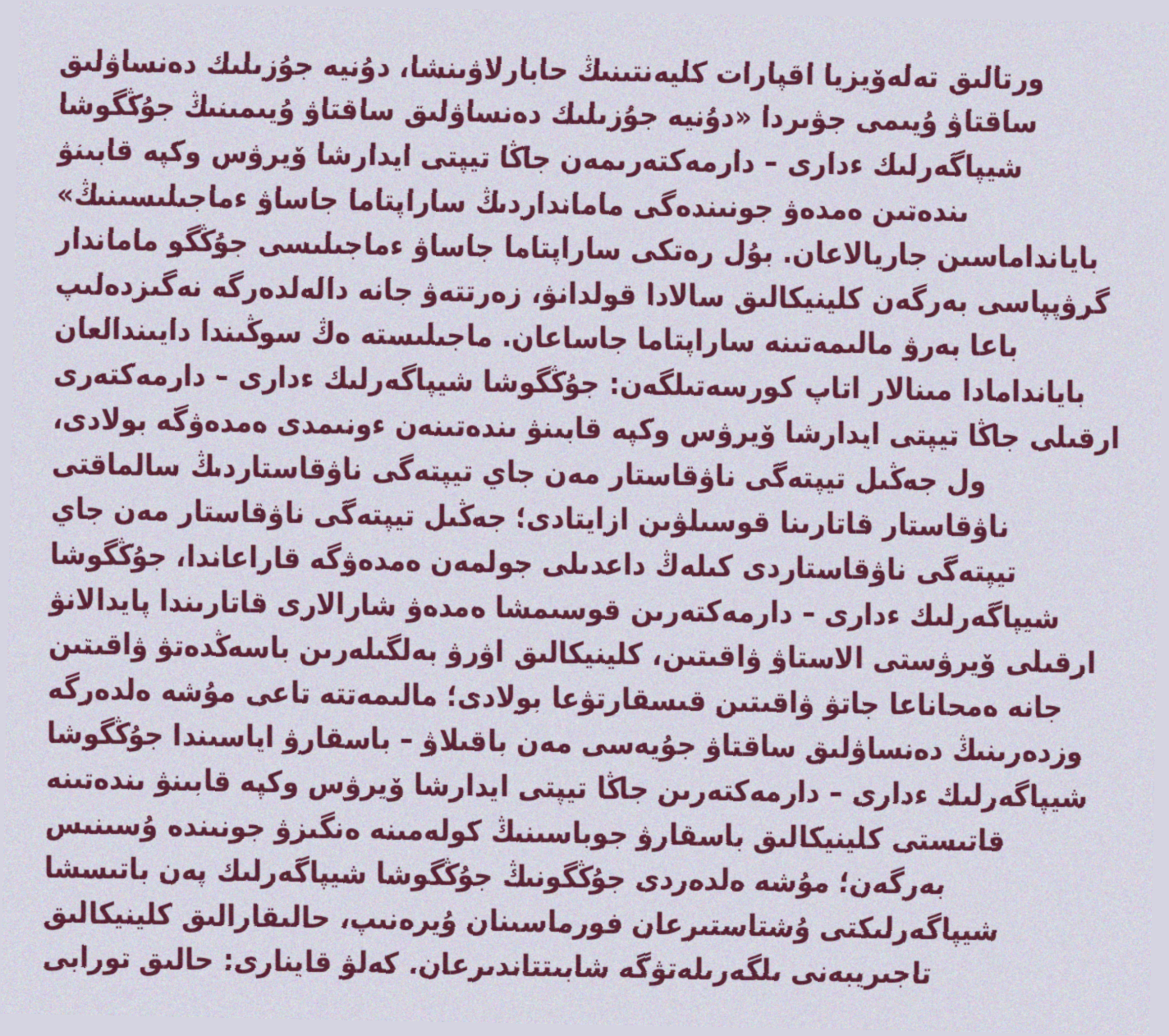}
    \caption{Example of an OCR task in the Kazakh Arabic script}
    \label{fig:arabicex}
\end{figure}

\begin{figure}
    \centering
    \includegraphics[width=1\linewidth]{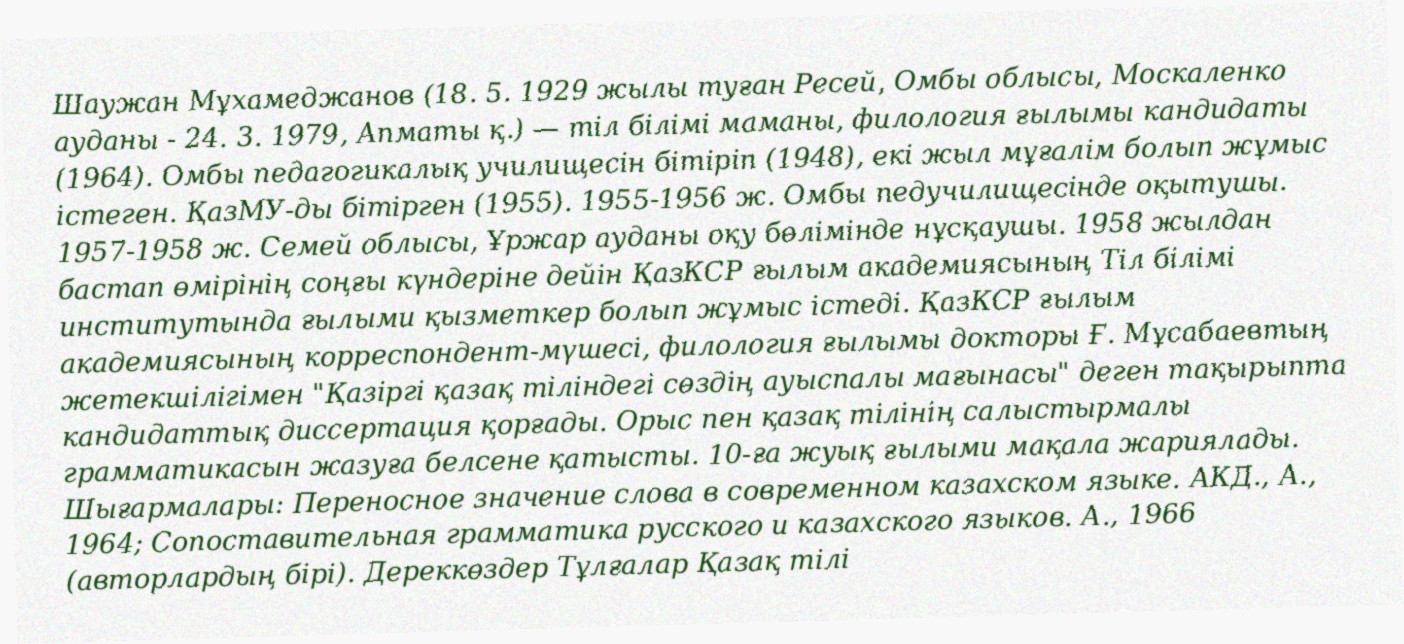}
    \caption{Example of an OCR task in the Kazakh Cyrillic script}
    \label{fig:cyrillicex}
\end{figure}

\begin{figure}
    \centering
    \includegraphics[width=1\linewidth]{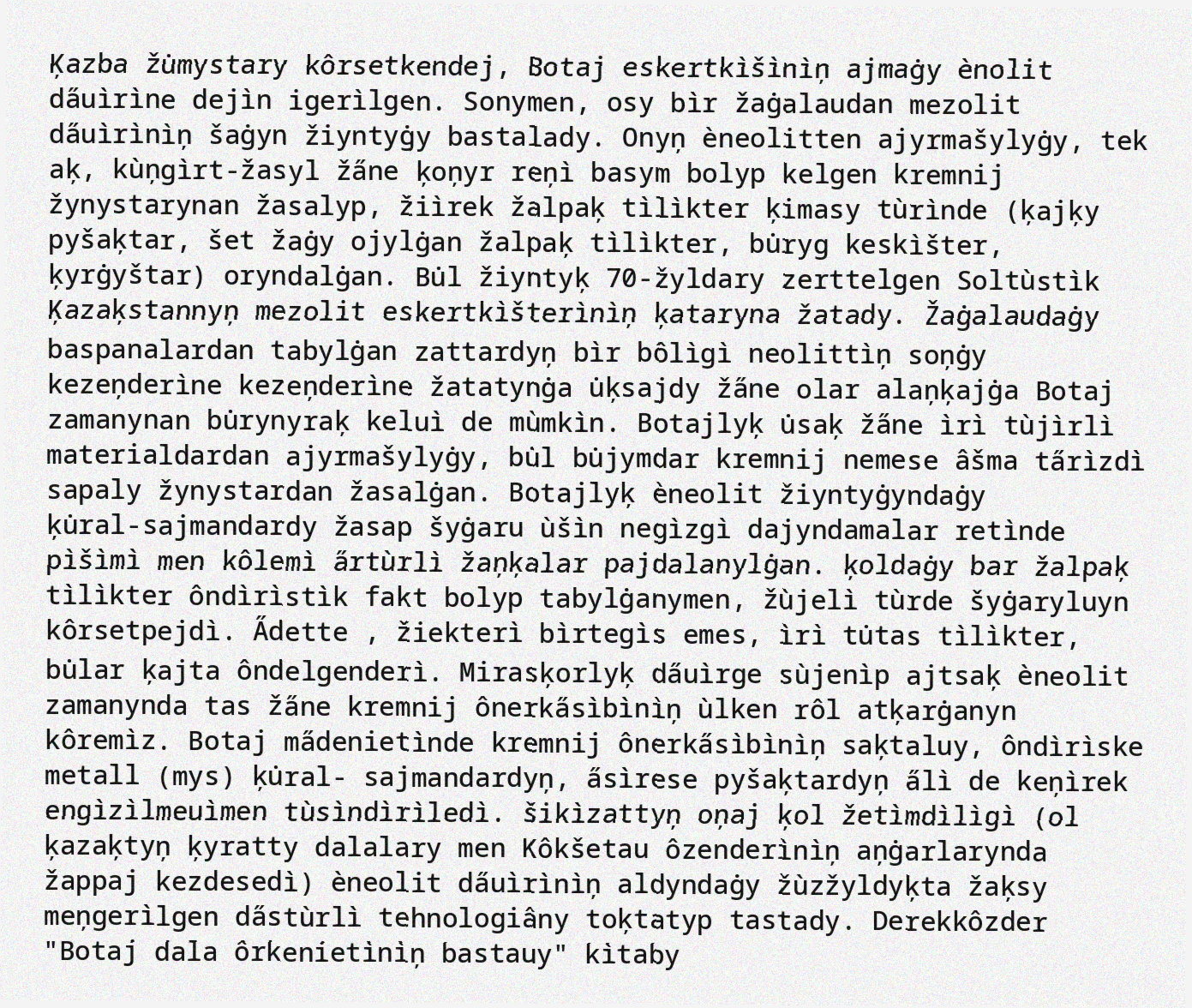}
    \caption{Example of an OCR task in the Kazakh Latin script}
    \label{fig:latinex}
\end{figure}

In order to make synthetic images representative of authentic OCR tasks, such as document scanning, we implement many variations across images, using the Pillow Python library. Examples of the benchmark and its variation can be seen in Figure \ref{fig:arabicex}, \ref{fig:cyrillicex}, and \ref{fig:latinex}.
\begin{itemize}
\item \textbf{Fonts:} We downloaded 2,491 fonts from Google Fonts and filtered all fonts for compatibility with Kazakh Cyrillic, Latin, and Arabic script. We found 135 fonts compatible with Arabic, 431 with Cyrillic, and 1376 with Latin. These fonts were randomly selected to create synthetic images. We also varied font size randomly from 24 to 56 across images.
\item \textbf{Noise:} We set the noise level to be random between 4 and 18, and a blur effect to be randomly between 0 and 0.8. Images are all rotated randomly between -3 and 3 degrees randomly.
\item \textbf{Color:} We designed two color palettes, each with 40 colors. One contained light background colors, and one contained darker text colors. When these colors were randomly selected, a minimum contrast ratio of 4.5 was used as a standard in all images. 
\end{itemize}

To foster reproducibility and encourage future work on low-resource Kazakh scripts, we release the full benchmark, which is available at \href{https://huggingface.co/datasets/henrygagnier/kazakh-ocr}{https://huggingface.co/datasets/henrygagnier/kazakh-ocr}. 
\subsection{Multimodal Large Language Models}
Multimodal large language models (MLLMs) preserve the reasoning and decision-making capabilities of large language models (LLMs), but also empower multimodal tasks, such as OCR \cite{zhang2024mmllmsrecentadvancesmultimodal}. We evaluated \texttt{gemma-3-12b-it} \cite{gemmateam2025gemma3technicalreport}, \texttt{Qwen2.5-VL-7B-Instruct} \cite{bai2025qwen25vltechnicalreport}, and \texttt{Llama-3.2-11B-Vision-Instruct} \cite{grattafiori2024llama3herdmodels} using a subset of the benchmark, with 200 images per script, or a total of 600 images. We chose parameter-efficient variants of recent multimodal models for cost and computational efficiency, to ensure that our methodology remains accessible under low-computational budgets, while still covering diverse model architectures in our evaluation. A temperature of 0.0 was used in all models with the prompt in Appendix \ref{sec:appendix}.
\subsection{Classical OCR Baseline}
We evaluate a classical OCR baseline using Tesseract as a baseline. We use the Tesseract OCR engine \cite{TessOverview} with its pretrained models on the Arabic, Latin, and Cyrillic scripts for each of the Kazakh scripts. No fine-tuning or adaptations were made, and the models were evaluated on an identical subset of 200 images per script. As Tesseract does not perform language identification, we only report character error rate and word error rate.
\section{Results}
We now look at the performance of the three models in the Arabic, Latin, and Cyrillic Kazakh scripts to find which scripts multimodal models experience challenges in (Table \ref{tab:ocrperformance}). In all models, character error rate (CER) and word error rate (WER) were best in Cyrillic script, and worst in Arabic script. Generally, Qwen performed the best across the three publicly available models, with a CER of 5.3\% in the Cyrillic script and a WER of 13.4\%. In the Latin and Arabic scripts, CER and WER are considerably worse, with a CER of 26.4\% and 35.5\% in the Latin and Arabic scripts, respectively. 

 All models are extremely unsuccessful at recognizing the Arabic script as Kazakh, with the Arabic script most commonly being misclassified as Arabic, Farsi, and Kurdish. The percent of samples correctly classified as Kazakh ranged from 0.0\% to 1.1\%. Qwen was highly successful at identifying the Cyrillic text as Kazakh, while also having the best OCR accuracy in the Cyrillic script. Gemma had an accuracy of 70.1\%, most commonly misclassifying Kazakh as Kyrgyz. Llama had an accuracy of 11.3\%, also most commonly misclassifying Kazakh as Kyrgyz. In the Latin script, Qwen was the most successful with an accuracy of 99.4\%, while Llama and Gemma had lower accuracies of 13.5\% to 31.1\%, with the Kazakh being most commonly misclassified as Kyrgyz and Tatar. Seen in Qwen and Gemma, models struggled in identifying low-resource scripts, while being successful with the Cyrillic script.

Across all three scripts, Tesseract achieves lower CERs than the evaluated MLLMs. This gap is the biggest in the Arabic script when CER is 15.0\% when using Tesseract, compared to 35.5\%–72.5\% when using the MLLMs. While there are large gaps in Arabic and Latin scripts, Qwen achieves a low CER similar to Tesseract. These results indicate that MLLMs, despite their general multimodal capabilities, do not match the OCR accuracy of traditional approaches on low-resource Kazakh scripts.
\vspace{-3mm}
\begin{table*}[ht]
\centering
\begin{tabular}{lcccc}
\toprule
\textbf{Model} & \textbf{Script} & \textbf{CER (\%)} & \textbf{WER (\%)} & \textbf{\% Kazakh} \\
\midrule
\multirow{3}{*}{Qwen2.5-VL-7B-Instruct} 
 & Arabic  & 35.5 & 90.0 & 1.1\% \\
 & Latin   & 26.4 & 57.4 & 99.4\% \\
 & Cyrillic& 5.3& 13.4 & 100.0\% \\
\midrule
\multirow{3}{*}{Llama-3.2-11B-Vision-Instruct} 
 & Arabic  & 37.3 & 90.2 & 0.0\% \\
 & Latin   & 21.6 & 66.3 & 31.1\% \\
 & Cyrillic& 18.2 & 49.8 & 11.3\% \\
\midrule
\multirow{3}{*}{Gemma-3-12B-it} 
 & Arabic  & 72.5 & 103.0& 0.0\% \\
 & Latin   & 31.0 & 79.2 & 13.5\% \\
 & Cyrillic& 25.7 & 42.9 & 70.1\% \\
 \midrule
 \multirow{3}{*}{Tesseract} 
 & Arabic  & 15.0 & 47.5 & —\\
 & Latin   & 11.4 & 51.6 & — \\
 & Cyrillic& 4.3 & 30.3 & — \\
\bottomrule
\end{tabular}
\caption{OCR Performance metrics of models across different scripts. CER = Character Error Rate, WER = Word Error Rate, and the percentage of samples correctly identified as Kazakh}
\label{tab:ocrperformance}
\end{table*}

\section{Discussion}
We construct KazakhOCR, the first OCR benchmark for low-resource Kazakh scripts. Evaluating multimodal large language models on the benchmark, we find significant disparities in OCR performance and language identification across the three scripts. The Cyrillic script, or the primary script in which Kazakh is written, achieved the best results across models, with Qwen2.5-VL-7B-Instruct achieving a CER of 5.3\% and a WER of 13.4\%. The Latin script had CERs ranging from 21.6 to 31.0\%, and the Arabic script had CERs ranging from 35.5\% to 72.5\%.

Comparing MLLMs with Tesseract reveals that traditional approaches have much greater accuracy than MLLMs, with lower CERs in all three scripts. The gap in performance is more pronounced in the Latin and Arabic scripts, demonstrating that current MLLM capabilities do not match OCR system capabilities for low-resource scripts. All MLLMs failed at identifying the Kazakh Arabic script, with correct identification rates of 0.0\% to 1.1\%. Most commonly misidentified as Arabic, Farsi, and Kurdish, this incapability of the model displays a lack in current MLLM understanding of low-resource Kazakh scripts. We theorize that this error is especially important in MLLM-based OCR because MLLMs often hallucinate \cite{he2025seeingbelievingmitigatingocr}, and when hallucinating in OCR for low-resource scripts, they may use words, characters, and diacritics found in languages that predict the text uses, leading to low OCR CER and WER. Qwen almost perfectly identified the Latin and Kazakh scripts, with accuracies of 99.4\% and 100.0\%, respectively, displaying that the Kazakh Arabic script is the least recognized by current MLLMs. These results are especially concerning for Kazakh speakers in China, Afghanistan, Pakistan, and Iran, where MLLMs do not adequately support the Kazakh Arabic script.

Future work with multimodal models for OCR should further experiment on prompt sensitivity and engineering for optimal OCR results. Work should also use authentic Latin script Kazakh text to properly represent cultural differences in the use of the script. Future LLMs and MLLMs should strive to support low-resource Kazakh scripts, considering the use of the Arabic script and the growing use of the Latin script. Future work should also use authentic low-resource Kazakh script texts, as opposed to synthetic benchmarks, when available. 

This study demonstrates that multimodal large language models have extreme disparities in processing low-resource Kazakh scripts, particularly for the Kazakh Arabic script. The KazakhOCR benchmark reveals significant gaps in current model capabilities for low-resource NLP, and highlights the need for inclusive training processes that support the Kazakh Latin and Arabic scripts. By making this benchmark publicly accessible, we aim to address these disparities in Kazakh NLP and encourage the development of more equitable multimodal models for the diverse Kazakh scripts.
\section{Conclusion}
This study synthetically constructs the first benchmark for low-resource Kazakh script OCR and identification. We evaluate Qwen2.5-VL-7B-Instruct, Llama-3.2-11B-Vision-Instruct, and Gemma-3-12B-it, three publicly available multimodal large language models, on our benchmark.

We find that all models are unsuccessful with Latin and Arabic script Kazakh OCR. Models have CERs of 35.5\% to 72.5\% for the Arabic script, and CERs of 26.4\% to 31.0\% for the Latin script. All models were unable to identify the Arabic script Kazakh as Kazakh, mistaking it for Arabic, Farsi, and Kurdish. Models were also largely unable to identify the Latin script. In all models, the best performance was on the Cyrillic script, with Qwen achieving a CER of 5.3\% and a WER of 13.4\%,

These findings display that low-resource Kazakh scripts are largely unsupported in MLLMs, with models failing at OCR and language identification, and demonstrate the necessity for the inclusion of the Kazakh Arabic and Kazakh Latin scripts in NLP.
\section*{Limitations}
Several limitations should be considered in this study. Due to the computational costs and time of MLLMs, we chose a sample of 200 images per script. This sample may not be large enough to represent the diversity in the Kazakh scripts. We also used transliterated Latin texts, which may have different uses and cultural differences from the Cyrillic script, although this has not been studied. This study also constructs a synthetic benchmark, due to the lack of real OCR data for low-resource Kazakh scripts, which may not capture all characteristics of real-world documents. 
\bibliography{ref}

@misc{Haq_Zhang_Khan_2025,
      title={PsOCR: Benchmarking Large Multimodal Models for Optical Character Recognition in Low-resource Pashto Language}, 
      author={Ijazul Haq and Yingjie Zhang and Irfan Ali Khan},
      year={2026},
      eprint={2505.10055},
      archivePrefix={arXiv},
      primaryClass={cs.CV},
      url={https://arxiv.org/abs/2505.10055}, 
}

@INPROCEEDINGS{953967,
  author={Margner, V. and Pechwitz, M.},
  booktitle={Proceedings of Sixth International Conference on Document Analysis and Recognition}, 
  title={Synthetic data for Arabic OCR system development}, 
  year={2001},
  volume={},
  number={},
  pages={1159-1163},
  doi={10.1109/ICDAR.2001.953967}}

@misc{saini2022ocrsyntheticbenchmarkdataset,
      title={OCR Synthetic Benchmark Dataset for Indic Languages}, 
      author={Naresh Saini and Promodh Pinto and Aravinth Bheemaraj and Deepak Kumar and Dhiraj Daga and Saurabh Yadav and Srihari Nagaraj},
      year={2022},
      eprint={2205.02543},
      archivePrefix={arXiv},
      primaryClass={cs.CV},
      url={https://arxiv.org/abs/2205.02543}, 
}

@misc{guan2024advancingpostocrcorrectioncomparative,
      title={Advancing Post-OCR Correction: A Comparative Study of Synthetic Data}, 
      author={Shuhao Guan and Derek Greene},
      year={2024},
      eprint={2408.02253},
      archivePrefix={arXiv},
      primaryClass={cs.CL},
      url={https://arxiv.org/abs/2408.02253}, 
}

@article{McCollum_Chen_2020, title={Kazakh}, volume={51}, url={http://dx.doi.org/10.1017/S0025100319000185}, DOI={10.1017/s0025100319000185}, abstractNote={<jats:p>Kazakh (ISO 639-3, kaz) is a Kipchak (Northwestern) Turkic language with approximately ten million speakers (Muhamedowa 2015). While the majority of Kazakh speakers live in the Republic of Kazakhstan, significant Kazakh-speaking populations exist throughout Central Asia. See Figure 1 for a map of the region. Kazakh spoken in Kazakhstan is described as having three or four dialects, but many researchers agree that differences between dialects are small and largely lexical (Kara 2002, Grenoble 2003, Muhamedowa 2015; see Amanzholov 1959 for more on Kazakh dialects).</jats:p>}, number={2}, journal={Journal of the International Phonetic Association}, publisher={Cambridge University Press (CUP)}, author={McCollum, Adam G. and Chen, Si}, year={2020}, month=feb, pages={276–298}, language={en} }

@article{honkasalo_temirbekova_2024,
  title        = {The Writing Reform and ‘Latinization’ of Written Kazakh: a Sociolinguistic Survey},
  author       = {Honkasalo, Sami and Temirbekova, Tansulu},
  journal      = {International Journal of Eurasian Linguistics},
  volume       = {6},
  number       = {1},
  pages        = {48--80},
  year         = {2024},
  doi          = {10.1163/25898833-20240056},
}

@article{Faizullah_Ayub_Hussain_Khan_2023, title={A Survey of OCR in Arabic Language: Applications, Techniques, and Challenges}, volume={13}, url={http://dx.doi.org/10.3390/app13074584}, DOI={10.3390/app13074584}, abstractNote={<jats:p>Optical character recognition (OCR) is the process of extracting handwritten or printed text from a scanned or printed image and converting it to a machine-readable form for further data processing, such as searching or editing. Automatic text extraction using OCR helps to digitize documents for improved productivity and accessibility and for preservation of historical documents. This paper provides a survey of the current state-of-the-art applications, techniques, and challenges in Arabic OCR. We present the existing methods for each step of the complete OCR process to identify the best-performing approach for improved results. This paper follows the keyword-search method for reviewing the articles related to Arabic OCR, including the backward and forward citations of the article. In addition to state-of-art techniques, this paper identifies research gaps and presents future directions for Arabic OCR.</jats:p>}, number={7}, journal={Applied Sciences}, publisher={MDPI AG}, author={Faizullah, Safiullah and Ayub, Muhammad Sohaib and Hussain, Sajid and Khan, Muhammad Asad}, year={2023}, month=apr, pages={4584}, language={en} }

@article{Razaque_Makezhanuly_Alimseitov_Kalpeyeva_Ayapbergenova_2024, title={Development of Handwritten Text Recognition system for the Kazakh Language}, volume={2}, url={https://ce.journal.satbayev.university/index.php/journal/article/view/1294}, DOI={10.51301/ce.2024.i4.01}, abstractNote={&amp;lt;p class=&amp;quot;Text&amp;quot;&amp;gt;&amp;lt;span lang=&amp;quot;EN-US&amp;quot;&amp;gt;The low digitalization of the Kazakh language is a problem that affects bureaucracy efficiency, the accessibility of literature, and education in the Kazakh language. This research introduces a modern approach to handwritten text recognition (HTR) for the Kaz akh language. It optimizes document flow and text mining, increases accessibility to Kazakh literature and historical resources, helps teachers in students’ essay scoring, and judges in decision -making. This solution optimizes operational processes in business, education, and government services. The state -of-the-art algorithms are integrated to achie ve improved accuracy and performance of text translation. HTR for the Kazakh language uses effective machine learning (ML) methods to create an HTR system specifically tuned for the Kazakh script. The se leverage features of Convolutional Neural Networks (CNN), Recurrent Neural Networks (RNN), image augmentation, transfer learning, and classic ML methods. HTR is implemented using Python programming language, O penCV, PyTorch, and Scikit- learn libraries. The system was trained on a large dataset of Kazakh handwritten text with different topics.&amp;lt;/span&amp;gt;&amp;lt;/p&amp;gt;}, number={4}, journal={Computing \& Engineering}, author={Razaque, A. and Makezhanuly, B. and Alimseitov, O. and Kalpeyeva, Zh. and Ayapbergenova, A.}, year={2024}, month={Dec.}, pages={1–7} }

@article{Toiganbayeva_Kasem_Abdimanap_Bostanbekov_Abdallah_Alimova_Nurseitov_2022, title={KOHTD: Kazakh offline handwritten text dataset}, volume={108}, url={http://dx.doi.org/10.1016/j.image.2022.116827}, DOI={10.1016/j.image.2022.116827}, journal={Signal Processing: Image Communication}, publisher={Elsevier BV}, author={Toiganbayeva, Nazgul and Kasem, Mahmoud and Abdimanap, Galymzhan and Bostanbekov, Kairat and Abdallah, Abdelrahman and Alimova, Anel and Nurseitov, Daniyar}, year={2022}, month=oct, pages={116827}, language={en} }

@article{Pirniyazova_Son_Kulzhan_2023, title={Recognition of latin letters of the kazakh alphabet based on a neural network}, volume={1}, url={https://ce.journal.satbayev.university/index.php/journal/article/view/1266}, DOI={10.51301/ce.2023.i3.07}, abstractNote={&amp;lt;p&amp;gt;This article discusses the recognition of Latin letters of the Kazakh alphabet based on a neural network using the Tensorflow framework. The paper provides an overview of existing recognition methods, including methods based on machine learning. To do this, a database of Kazakh letters of the Latin alphabet is being created using the Turtle graphic library. A computational algorithm for recognizing Kazakh letters of the Latin alphabet has been developed. The algorithm uses operations on labels entered for letters to predict the target variable. The model is trained on the prepared data and then evaluated based on its performance on the validation dataset. The overall accuracy of the model is approximately 93.90%. This means that about 93.90% of the model’s predictions were correct. The paper uses two metrics for the accuracy Precision and completeness Recall classes. These metrics show whether the models work well by class, show information about the performance of the model. The experiments have shown that the proposed algorithm provides high accuracy of recognition of letters of the Latin Kazakh alphabet. The results are illustrated graphically. In the discussion of the results, a multiclass ROC curve is presented, which is a graphical representation of the performance of the classification model at all classification thresholds. The performance of the model as a whole indicates the high performance of the classification model. The algorithm proposed in the article for recognizing Latin letters of the Kazakh alphabet can be used in various applications, such as optical character recognition (OCR) systems, automated verification of entered texts using mobile devices.&amp;lt;/p&amp;gt;}, number={3}, journal={Computing \& Engineering}, author={Pirniyazova, P. and Son, E.Yu. and Kulzhan, D.Zh.}, year={2023}, month={Sep.}, pages={36–41} }

@article{Yeleussinov_Amirgaliyev_Cherikbayeva_2023, title={Improving OCR Accuracy for Kazakh Handwriting Recognition Using GAN Models}, volume={13}, url={http://dx.doi.org/10.3390/app13095677}, DOI={10.3390/app13095677}, abstractNote={<jats:p>This paper aims to increase the accuracy of Kazakh handwriting text recognition (KHTR) using the generative adversarial network (GAN), where a handwriting word image generator and an image quality discriminator are constructed. In order to obtain a high-quality image of handwritten text, the multiple losses are intended to encourage the generator to learn the structural properties of the texts. In this case, the quality discriminator is trained on the basis of the relativistic loss function. Based on the proposed structure, the resulting document images not only preserve texture details but also generate different writer styles, which provides better OCR performance in public databases. With a self-created dataset, images of different types of handwriting styles were obtained, which will be used when training the network. The proposed approach allows for a character error rate (CER) of 11.15% and a word error rate (WER) of 25.65%.</jats:p>}, number={9}, journal={Applied Sciences}, publisher={MDPI AG}, author={Yeleussinov, Arman and Amirgaliyev, Yedilkhan and Cherikbayeva, Lyailya}, year={2023}, month=may, pages={5677}, language={en} }

@inproceedings{zhang-etal-2024-mc2,
    title = "{MC}$^2$: Towards Transparent and Culturally-Aware {NLP} for Minority Languages in {C}hina",
    author = "Zhang, Chen  and
      Tao, Mingxu  and
      Huang, Quzhe  and
      Lin, Jiuheng  and
      Chen, Zhibin  and
      Feng, Yansong",
    editor = "Ku, Lun-Wei  and
      Martins, Andre  and
      Srikumar, Vivek",
    booktitle = "Proceedings of the 62nd Annual Meeting of the Association for Computational Linguistics (Volume 1: Long Papers)",
    month = aug,
    year = "2024",
    address = "Bangkok, Thailand",
    publisher = "Association for Computational Linguistics",
    url = "https://aclanthology.org/2024.acl-long.479/",
    doi = "10.18653/v1/2024.acl-long.479",
    pages = "8832--8850"
}

@InProceedings{10.1007/978-3-030-60276-5_63,
author="Yessenbayev, Zhandos
and Kozhirbayev, Zhanibek
and Makazhanov, Aibek",
editor="Karpov, Alexey
and Potapova, Rodmonga",
title="KazNLP: A Pipeline for Automated Processing of Texts Written in Kazakh Language",
booktitle="Speech and Computer",
year="2020",
publisher="Springer International Publishing",
address="Cham",
pages="657--666",
isbn="978-3-030-60276-5"
}

@misc{zhang2024mmllmsrecentadvancesmultimodal,
      title={MM-LLMs: Recent Advances in MultiModal Large Language Models}, 
      author={Duzhen Zhang and Yahan Yu and Jiahua Dong and Chenxing Li and Dan Su and Chenhui Chu and Dong Yu},
      year={2024},
      eprint={2401.13601},
      archivePrefix={arXiv},
      primaryClass={cs.CL},
      url={https://arxiv.org/abs/2401.13601}, 
}

@misc{bai2025qwen25vltechnicalreport,
      title={Qwen2.5-VL Technical Report}, 
      author={Shuai Bai and Keqin Chen and Xuejing Liu and Jialin Wang and Wenbin Ge and Sibo Song and Kai Dang and Peng Wang and Shijie Wang and Jun Tang and Humen Zhong and Yuanzhi Zhu and Mingkun Yang and Zhaohai Li and Jianqiang Wan and Pengfei Wang and Wei Ding and Zheren Fu and Yiheng Xu and Jiabo Ye and Xi Zhang and Tianbao Xie and Zesen Cheng and Hang Zhang and Zhibo Yang and Haiyang Xu and Junyang Lin},
      year={2025},
      eprint={2502.13923},
      archivePrefix={arXiv},
      primaryClass={cs.CV},
      url={https://arxiv.org/abs/2502.13923}, 
}

@misc{he2025seeingbelievingmitigatingocr,
      title={Seeing is Believing? Mitigating OCR Hallucinations in Multimodal Large Language Models}, 
      author={Zhentao He and Can Zhang and Ziheng Wu and Zhenghao Chen and Yufei Zhan and Yifan Li and Zhao Zhang and Xian Wang and Minghui Qiu},
      year={2025},
      eprint={2506.20168},
      archivePrefix={arXiv},
      primaryClass={cs.CV},
      url={https://arxiv.org/abs/2506.20168}, 
}

@misc{gemmateam2025gemma3technicalreport,
      title={Gemma 3 Technical Report}, 
      author={Gemma Team and Aishwarya Kamath and Johan Ferret and Shreya Pathak and Nino Vieillard and Ramona Merhej and Sarah Perrin and Tatiana Matejovicova and Alexandre Ramé and Morgane Rivière and Louis Rouillard and Thomas Mesnard and Geoffrey Cideron and Jean-bastien Grill and Sabela Ramos and Edouard Yvinec and Michelle Casbon and Etienne Pot and Ivo Penchev and Gaël Liu and Francesco Visin and Kathleen Kenealy and Lucas Beyer and Xiaohai Zhai and Anton Tsitsulin and Robert Busa-Fekete and Alex Feng and Noveen Sachdeva and Benjamin Coleman and Yi Gao and Basil Mustafa and Iain Barr and Emilio Parisotto and David Tian and Matan Eyal and Colin Cherry and Jan-Thorsten Peter and Danila Sinopalnikov and Surya Bhupatiraju and Rishabh Agarwal and Mehran Kazemi and Dan Malkin and Ravin Kumar and David Vilar and Idan Brusilovsky and Jiaming Luo and Andreas Steiner and Abe Friesen and Abhanshu Sharma and Abheesht Sharma and Adi Mayrav Gilady and Adrian Goedeckemeyer and Alaa Saade and Alex Feng and Alexander Kolesnikov and Alexei Bendebury and Alvin Abdagic and Amit Vadi and András György and André Susano Pinto and Anil Das and Ankur Bapna and Antoine Miech and Antoine Yang and Antonia Paterson and Ashish Shenoy and Ayan Chakrabarti and Bilal Piot and Bo Wu and Bobak Shahriari and Bryce Petrini and Charlie Chen and Charline Le Lan and Christopher A. Choquette-Choo and CJ Carey and Cormac Brick and Daniel Deutsch and Danielle Eisenbud and Dee Cattle and Derek Cheng and Dimitris Paparas and Divyashree Shivakumar Sreepathihalli and Doug Reid and Dustin Tran and Dustin Zelle and Eric Noland and Erwin Huizenga and Eugene Kharitonov and Frederick Liu and Gagik Amirkhanyan and Glenn Cameron and Hadi Hashemi and Hanna Klimczak-Plucińska and Harman Singh and Harsh Mehta and Harshal Tushar Lehri and Hussein Hazimeh and Ian Ballantyne and Idan Szpektor and Ivan Nardini and Jean Pouget-Abadie and Jetha Chan and Joe Stanton and John Wieting and Jonathan Lai and Jordi Orbay and Joseph Fernandez and Josh Newlan and Ju-yeong Ji and Jyotinder Singh and Kat Black and Kathy Yu and Kevin Hui and Kiran Vodrahalli and Klaus Greff and Linhai Qiu and Marcella Valentine and Marina Coelho and Marvin Ritter and Matt Hoffman and Matthew Watson and Mayank Chaturvedi and Michael Moynihan and Min Ma and Nabila Babar and Natasha Noy and Nathan Byrd and Nick Roy and Nikola Momchev and Nilay Chauhan and Noveen Sachdeva and Oskar Bunyan and Pankil Botarda and Paul Caron and Paul Kishan Rubenstein and Phil Culliton and Philipp Schmid and Pier Giuseppe Sessa and Pingmei Xu and Piotr Stanczyk and Pouya Tafti and Rakesh Shivanna and Renjie Wu and Renke Pan and Reza Rokni and Rob Willoughby and Rohith Vallu and Ryan Mullins and Sammy Jerome and Sara Smoot and Sertan Girgin and Shariq Iqbal and Shashir Reddy and Shruti Sheth and Siim Põder and Sijal Bhatnagar and Sindhu Raghuram Panyam and Sivan Eiger and Susan Zhang and Tianqi Liu and Trevor Yacovone and Tyler Liechty and Uday Kalra and Utku Evci and Vedant Misra and Vincent Roseberry and Vlad Feinberg and Vlad Kolesnikov and Woohyun Han and Woosuk Kwon and Xi Chen and Yinlam Chow and Yuvein Zhu and Zichuan Wei and Zoltan Egyed and Victor Cotruta and Minh Giang and Phoebe Kirk and Anand Rao and Kat Black and Nabila Babar and Jessica Lo and Erica Moreira and Luiz Gustavo Martins and Omar Sanseviero and Lucas Gonzalez and Zach Gleicher and Tris Warkentin and Vahab Mirrokni and Evan Senter and Eli Collins and Joelle Barral and Zoubin Ghahramani and Raia Hadsell and Yossi Matias and D. Sculley and Slav Petrov and Noah Fiedel and Noam Shazeer and Oriol Vinyals and Jeff Dean and Demis Hassabis and Koray Kavukcuoglu and Clement Farabet and Elena Buchatskaya and Jean-Baptiste Alayrac and Rohan Anil and Dmitry and Lepikhin and Sebastian Borgeaud and Olivier Bachem and Armand Joulin and Alek Andreev and Cassidy Hardin and Robert Dadashi and Léonard Hussenot},
      year={2025},
      eprint={2503.19786},
      archivePrefix={arXiv},
      primaryClass={cs.CL},
      url={https://arxiv.org/abs/2503.19786}, 
}

@inproceedings{TessOverview,
  author = {Ray Smith},
  title = {An Overview of the Tesseract OCR Engine},
  booktitle = {ICDAR '07: Proceedings of the Ninth International Conference on Document Analysis and Recognition},
  url = {https://storage.googleapis.com/pub-tools-public-publication-data/pdf/33418.pdf},
  year = {2007},
  isbn = {0-7695-2822-8},
  pages = {629--633},
  publisher = {IEEE Computer Society},
  address = {Washington, DC, USA},
}

\appendix
\section{MLLM Prompt}
\label{sec:appendix}
We used the following prompt for all MLLM evaluations:
\begin{quote}
\small
\texttt{You are an expert OCR system.}

\texttt{Read and transcribe all text visible in the image.}

\texttt{Preserve exact formatting, spacing, and punctuation.}

\texttt{Identify the language using ISO 639 codes.}

\texttt{Return only valid minified JSON in this exact format:}

\texttt{\{"language":"iso\_code","text":"exact transcription"\}}
\end{quote}

\end{document}